# Multimodal Emotion Recognition for One-Minute-Gradual Emotion Challenge

Ziqi Zheng          Chenjie Cao          Xingwei Chen          Guoqiang Xu

Ping An, Gamma Lab, Shanghai, China

{zhengziqi365, caochenjie948, chenxingwei068, xuguoqiang371}@pingan.com.cn

**ABSTRACT**

The continuous dimensional emotion modelled by arousal and valence can depict complex changes of emotions. In this paper, we present our works on arousal and valence predictions for One-Minute-Gradual (OMG) Emotion Challenge. Multimodal representations are first extracted from videos using a variety of acoustic, video and textual models and support vector machine (SVM) is then used for fusion of multimodal signals to make final predictions. Our solution achieves Concordant Correlation Coefficient (CCC) scores of 0.397 and 0.520 on arousal and valence respectively for the validation dataset, which outperforms the baseline systems with the best CCC scores of 0.15 and 0.23 on arousal and valence by a large margin.

## 1 INTRODUCTION

Facial expression plays vital roles in human communications [1]. Automatically human facial expression recognition (FER) helps improve human-computer interactions [2]. Classification of six prototypical facial expression—anger disgust, fear, happiness, sadness and surprise—are of most interests [3, 4]. However, various human facial expression can hardly be covered by the basic six emotions. The dimensional model of emotion [5] encodes intensity changes of emotion, including two continuous valued variables, arousal and valence, which can model subtle, complicated and continuous affective emotion states. Many studies start to focus on the modelling the arousal and valence emotion recently [6, 7].

The video-based OMG Emotion Challenge asks the competitors to develop methods to automatically judge the arousal and valence values for videos selected from Youtube. The OMG Emotion dataset contains around 420 videos, each video is split into utterances. The utterances are then annotated with arousal and valence scores by at least five independent persons [8]. This large and well-annotated dataset provides valuable footstones for automatically expression recognition research.

## 2 PROPOSED METHODS

### 2.1 Acoustic Feature Extraction

**OpenSmile Feature:** The acoustic representation of a video clip is generated using hand-crafted features. We first use OpenSmile [9] to extract 6552 audio features based on the whole audio file and then use XGBoost [10] for supervised feature selection. During the feature selection process, several decision tree models vote for the importance of acoustic features in classifying emotions. Finally, 256 audio features with the highest importance scores are obtained. We find that the skewness of the smoothed $6^{th}$ Mel-frequency cepstral coefficients, $1^{st}$ quartile of $1^{st}$ order delta coefficients of the smoothed voicing probability computed from the autocorrelation function, $2^{nd}$ quartile of $2^{nd}$ order delta coefficients of the smoothed $2^{nd}$ Mel-frequency cepstral coefficients are of great importance in predicting emotion states.

**SoundNet Feature:** We also explored deep learning models to extract the acoustic representation. The model we use is SoundNet with 8 convolution layers [11]. As a 1D convolution network, SoundNet can extract local audio features effectively [11]. When fusing with other modalities, it can improve the generalization performance of the fused model in our experiments. The SoundNet is trained with the 1D acoustic amplitude of each time point collected from the official audio datasets directly without any pre-training. The length of each sample is padded with zero to 600000 which is two times of the average length. The optimizer we used is SGD with momentum 0.5, learning rate 0.003 of 0.001 decay in each step, and the batch size is fixed at 64. Finally, 256D features are extract from pool5 of the SoundNet after about 200 epochs.

## 2.2 Visual Feature Extraction

To focus on facial expression, we utilize SeetaFace [12] to detect and crop human faces. The detector is actually a funnel-structure model which comprises a fast LAB cascade classifier and two MLP cascades.

Deep convolutional neural networks (CNN) are driving advances in multiple computer vision tasks since "AlexNet" [13] won the 2012 ImageNet competition [14]. CNN have also been successfully used to explore facial expression recognition [4, 15, 16] as CNN can reach excellent representations of image features. Classic deep CNN structures such as VGG [17] and GoogleNet [18], which were originally developed to solve image classification problems, have now been widely transferred to different kinds of computer vision tasks such as object detection [19, 20] and emotion recognitions [16].

Here we explored VGG, GoogleNet and one of the best performance FER networks, HoloNet [4] for facial emotion feature representations. We extracted both global average pooling (GAP) features after the last max pool layer and the fully connected (FC) layer features. When evaluate these representations on the OMG Emotion Challenge's validation datasets, global average pooling features of VGG16 outperforms other structures on both arousal and valence (Table 1). We also tried deeper CNN structures such as ResNet [21] and DenseNet [22], which exhibited poor performance on the validation datasets. VGG GAP representations were used for afterwards contextual feature generation.

**Table 1. Performance of different network representations**

| Nets | CCC | |
|---|---|---|
| | Arousal | Valence |
| VGG GAP | 0.169 | 0.420 |
| VGG FC | 0.138 | 0.393 |
| GoogleNet | 0.142 | 0.362 |
| HoloNet | 0.130 | 0.259 |

LSTM network [23] were always used to manipulate sequential data, and it also been widely used in video based emotion recognition [16, 24]. Another efficient approach for sequential data is the 1D-CNN, which have been successfully used in natural language processing [25, 26]. In the OMG Emotion datasets, each utterance contains several sequential frames of images, we used both LSTM [23] and concatenation of multi 1D-CNN with different filter size to employ the contextual information.

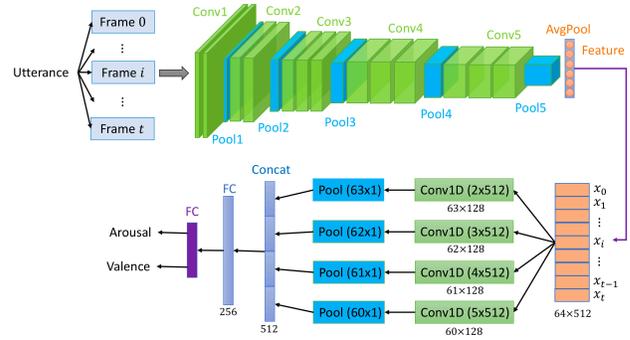

**Figure 1. Structure of VGG16 + 1DCNN network**

In the VGG16 + 1D-CNN model (Figure 1), representations of each frame in the utterance were extracted with VGG16 network (without FC layers) followed by global average pooling. Then 512-dimensional (512D) representations were concatenated across frames forming the 64×512 inputs for 1D-CNN network (utterance with less than 64 frames were padded with zeros). Four kinds of 1D convolutions (with filter size 2×512, 3×512, 4×512, 5×512, respectively) followed by global max pooling were used to extract the features. After concatenating the features of the four convolution layers, two FC layers were used to predict the continuous value of arousal and valence.

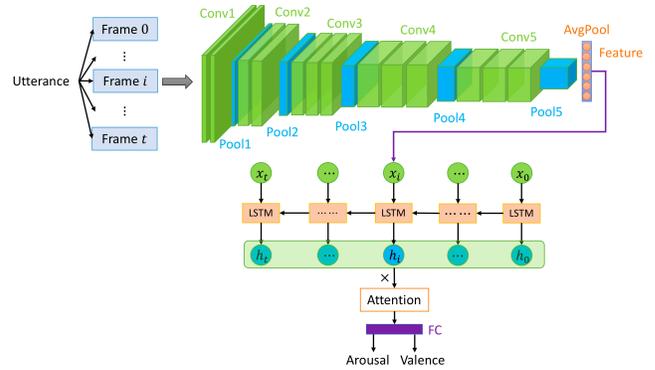

**Figure 2. Structure of VGG16 + LSTM network**

Similar with VGG16 + 1D-CNN model, the VGG16 + LSTM (Figure 2) also used VGG16 to extract feature representations. Then LSTM layer with 256 hidden cells were used to encode the input representations. We also adapted the attention model from [27] to weighting for the hidden cells across different frames. Assume the encoded hidden states across frames were denoted with $H = (h_0, h_1, \cdots, h_t)$, then the attention vector can be computed as:
$$V = tanh(HW + b)$$
$$a = softmax(Vu)$$
Then $a$ denotes the weight vector of the frames.

### 2.3 Text Feature Extraction

Each utterance contains a corresponding sentence in English. We use the twitter skip-gram model from gensim [28] with about 3 million 400-dimentional embedding word vectors as the pre-trained Word2Vector [29] model. Then, bidirectional GRU (Bi-GRU) [30] and Multi-Head Attention (MHA) [31] are used to encode these embedding features. As the textual features perform badly in arousal, we only apply them on predicting valence. The performances of Bi-GRU and MHA in validation dataset are shown in Table 2.

**Bi-GRU:** We use the bidirectional GRU (Bi-GRU) [30] with concatenating and a fully connected layer of 256 units with ReLU activation to extract the textual feature. The optimizer of Bi-GRU is Adam with learning rate fixed in 0.001, beta1=0.9 and beta2=0.98. However, there is no obvious advantages of Bi-GRU compared with MHA in Table 2. Since Bi-GRU takes too long to train, we give up on this model in the final fusion.

**MHA:** The Multi-Head Attention (MHA) is proposed by google in 2017 [31]. MHA can solve the time sequence problem only by the dot-product attention without any other CNN and RNN [31]. So, MHA can be trained much faster while the complexity is lower compared with other methods. We train the MHA with 8 parallel attention layers of 64 units and extract the feature from the subsequent 256D fully connected layer with ReLU. The optimizer of MHA is the SGD with momentum 0.5, and the learning rate is 0.005 with 0.001 decay for each step. Benefited from the simplify of MHA, the batch size can be enlarged to 64 to get better generalization.

**Table 2. Performance of Bi-GRU and MHA on valence CCC of validation**

| Textual methods | Valence CCC |
|---|---|
| Bi-GRU | 0.213 |
| MHA | 0.222 |

### 2.4 Multi-modal Feature Fusion

In order to take advantage of facial expression, speech and text, we decide to fuse features over different modalities. However, the direct fusion of features right extracted from raw data only brings slight improvement. As we know, high-level features can represent the data attributes better. Hence, we concatenate visual, acoustic and textual features in deep hierarchy and pass them into an SVM classifier to get multi-modal predictions (Figure 3).

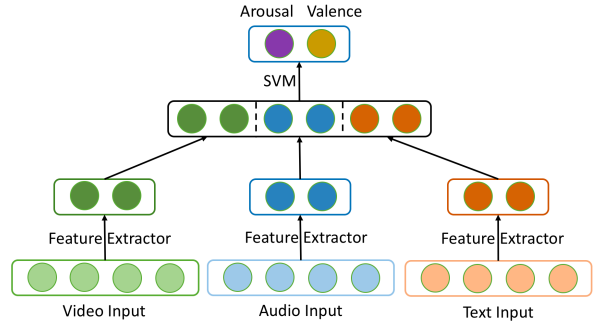

**Figure 3. Fusions of visual, acoustic and text features**

## 3 OPTIMIZATION STRATEGIES

Our target is to reduce the gap between predictions and ground truth. Specifically, we strive to find an optimum along the hypersurface of objective function. Furthermore, the solution is supposed to generalize well on unseen examples. We test a lot of strategies and analyze them in this section.

In all of our experiments, we keep every component and parameter same except for the part we are investigating.

### 3.1 Data Augmentation

In OMG database, we note that the duration of video clips vary a lot, from 2 seconds to one minute. In order to handle relatively long videos, we down-sampled the image sequences. The sampling rate is one every five frames in our experiments.

**Table 3. Performance of different data augmentations**

| Augmentation | CCC | |
|---|---|---|
| | Arousal | Valence |
| No | 0.169 | -0.420 |
| SSA | 0.196 | 0.443 |
| CSA | 0.129 | 0.447 |

However, the rest of frames are totally discarded. It is definitely harmful to emotion prediction. We apply two sequential data augmentation strategies in the training stage to avoid loss of information. The first one is Sample Sequential Augmentation (SSA) which means we select one frame randomly from every 5 frames. The other one is called Chunk Sequential Augmentation (CSA). We truncate a fixed length of consecutive frames ran-domly for each utterance. Table 3 states that CSA improves the performance on valence prediction whereas hurts the arousal prediction. The SSA can achieve better results on both arousal and valence predictions. Apparently as the training goes on, all the frames should be sampled and trained. Then we make sure we use all data source.

### 3.2 Loss Function

An appropriate loss function plays a big role in producing optimal results. Different loss functions can guide neural networks through different learning routes. Since this is a regression task, we firstly choose the Mean Square Error (MSE) and Mean Absolute Error (MAE) loss. In addition, we also explored the loss function incorporating batch-wise Concordance Correlation Coefficient (CCC) scores.

**Table 4. Performance of different loss functions**

| Loss | CCC | |
|---|---|---|
| | Arousal | Valence |
| MSE | 0.167 | 0.373 |
| MAE | 0.169 | 0.420 |
| CCC + MSE | 0.134 | 0.419 |
| CCC + MAE | 0.198 | 0.381 |

As illustrated in Table 4, MAE is a better objective than MSE for optimizing CCC between predictions and ground truths. Batch-wise CCC loss is able to help model predict more accurate arousal values. But we did not adopt CCC loss due to the extreme sensitivity to its proportion in loss. It is hard to evaluate a suitable proportion for non-labelled data.

### 3.3 Independent vs. Multi-task Learning

Table 5 lists the Pearson correlation between arousal and valence on training data and validation data. Obviously, there is some hidden relation between them. Therefore, we train the arousal and valence values jointly, i.e., via multi-task learning. The results in Table 5 show that multi-task learning can improve the performance of both arousal and valence predictions.

**Table 5. Performance on independent and multi-task learning**

| Learning Scheme | CCC | |
|---|---|---|
| | Arousal | Valence |
| Independent | 0.138 | 0.393 |
| Multi-task | 0.169 | 0.420 |

### 3.4 Fixed vs. Trainable FNN

We tested two training strategies for the visual model. One is to first extract frame-level representations using the VGG model and then train a sequential model while keeping frame-level representation vectors fixed. The other one is to train the VGG and sequential models together. The former training strategy can achieve faster convergence and more stable performance, while the latter offers larger capacity. Table 6 shows that full-network training leads to a higher CCC score of arousal predictions, while it performs poorly for valence predictions.

**Table 6. Performance with different training strategies**

| Training Strategies | CCC | |
|---|---|---|
| | Arousal | Valence |
| Fixed CNN | 0.169 | 0.420 |
| Fully Training | 0.251 | 0.348 |

## 4 EXPERIMENTS

### 4.1 Single-Modal Results

We first tested single-modal models. In each modal, representation vectors are extracted and then passed to a SVM classifier with RBF kernel. The kernel coefficient is fixed in $\sigma = 1/n_{feature}$. The best penalty parameter C is selected by grid-search according to a 5-folds cross validation. Results of each single model are shown in Table 7. As shown in Table 7, VidModel1 (FixedVGG16 + LSTM + Attention) and VidModel2 (FixedVGG16 + 1D-CNN) achieve the best result of arousal and valence respectively. Furthermore, the Pearson Correlation Coefficient (PCC) of these CCC results is shown in Figure 4. Figure 4 indicates that models from the same source earn higher correlations such as VidModel1 and VidModel2, while the correlations between the models from different modalities are comparatively lower.

**Table 7. Performance of single modal trained in 5-folds cross validation**

| Single Modal | CCC | |
|---|---|---|
| | Arousal | Valence |
| VisModel1 (FixedVGG16 + LSTM + Attention) | 0.307 | 0.520 |
| VisModel2 (FixedVGG16 + 1D-CNN) | 0.351 | 0.480 |
| AudModel1 (SoundNet) | 0.289 | 0.336 |
| AudModel2 (OpenSmile) | 0.342 | 0.395 |
| TextModel (Multi-Head Attention) | 0.103 | 0.234 |

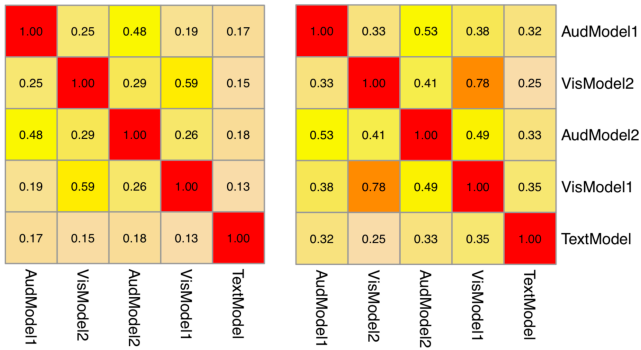

**Figure 4. PCC matrix of the results from Table 7**

## 5.2 Multi-Modal Results

Fusion is performed across different modalities. Acoustic, visual and textual features from VidModel1, VidModel2, AudModel1, AudModel2, TextModel are first extracted and then concentrated in a variety of combinations. Then, SVM with RBF kernel works as the final classifier for these concatenated features. In Table 8, results with good performance according to the 5-folds cross validation are shown.

**Table 8. Performance of multi modal trained in 5-folds cross validation**

| Multi Modal | CCC | | |
|---|---|---|---|
| | Arousal | Valence | Mean |
| VisModel1 + VisModel2 +AudModel1+AudModel2 | 0.392 | 0.515 | 0.454 |
| VisModel2 + AudModel2 | 0.397 | 0.490 | 0.443 |
| VisModel2 + AudModel1 + AudModel2 | 0.395 | 0.490 | 0.443 |
| VisModel1 + VisModel2 + AudModel1 | 0.360 | 0.516 | 0.438 |
| VisModel1 + TextModel | 0.305 | 0.511 | 0.408 |

## 5  CONCLUSIONS

In this paper, we present our models of arousal and valence predictions for the OMG Emotion Challenge. We explored various hand-crafted and deep learning models for generating acoustic, visual and textual representations. We also tested different model hyper-parameters and optimization strategies for better generalizations. Finally, the combination of hand-crafted and deep learned acoustic features and visual features learned from VGG16 followed by LSTM and 1DCNN models achieve the best CCC scores of 0.397 and 0.520 for arousal and valence respectively, which outperform the baseline 0.15 and 0.23.